\documentclass[letterpaper, 10 pt, conference]{ieeeconf}  % Comment this line out if you need a4paper

\IEEEoverridecommandlockouts                              % This command is only needed if you want to use the \thanks command

\usepackage[utf8]{inputenc} % allow utf-8 input
\usepackage[T1]{fontenc}    % use 8-bit T1 fonts
\usepackage{booktabs}       % professional-quality tables
\usepackage{amsfonts}       % blackboard math symbols
\usepackage{nicefrac}       % compact symbols for 1/2, etc.
\usepackage{microtype}      % microtypography
\usepackage{graphicx}       % graphics
\usepackage{amsmath}
\usepackage{mathtools} % coloneqq 
\usepackage{bm}
\usepackage{algorithm}
\usepackage{algpseudocode}
\usepackage{comment} 
\usepackage[font=normalsize,labelfont=bf]{caption}
\usepackage{siunitx}
\usepackage{cuted}
\usepackage{caption}

\usepackage[nocompress]{cite}

\makeatletter
\let\NAT@parse\undefined
\makeatother
\usepackage[dvipsnames,svgnames]{xcolor}
\usepackage{url}
\usepackage{hyperref}
\hypersetup{%
  colorlinks=true,%
  linkcolor={red!50!black!80!yellow},
  citecolor={blue!80!black},
  urlcolor={blue!80!black},
  bookmarksnumbered=true,
  bookmarksopen=true}

\usepackage{listings}
\definecolor{backcolour}{rgb}{0.98,0.98,0.93}

\newcommand{\listingsttfamily}{\fontfamily{IBMPlexMono-TLF}\scriptsize}
\lstdefinestyle{prettycode}{
  basicstyle=\listingsttfamily,
  backgroundcolor=\color{backcolour},
  aboveskip={0.9\baselineskip},               
  keepspaces=true,
}
\newcommand{\allen}[1]{{\normalsize{\color{blue}{[\textbf{AR: }#1]}}}}

\newcommand{\asher}[1]{{\normalsize{\color{purple}{[\textbf{AH: }#1]}}}}

\newcommand{\byovla}{BYO\textsl{VLA}}

\title{\LARGE \bf
Run-time Observation Interventions \\ Make Vision-Language-Action Models More Visually Robust
}

\author{Asher J. Hancock$^{1}$, Allen Z. Ren$^{1}$, and Anirudha Majumdar$^{1}$% <-this % stops a space
\thanks{*This work was partially supported by the NSF CAREER Award [\#2044149] and the Office of Naval Research [N00014-23-1-2148]. Asher Hancock was supported by the National Science Foundation Graduate Research Fellowship Program under Grant No. DGE-2146755.}% <-this % stops a space
\thanks{$^{1}$Dept. of Mechanical \& Aerospace Engineering, Princeton University. Contact: {\tt\small ajhancock@princeton.edu}.}%
}

\begin{document}
\maketitle
\thispagestyle{empty}
\pagestyle{empty}
%%%%%%%%%%%%%%%%%%%%%%%%%%%%%%%%%%%%%%%%%%%%%%%%%%%%%%%%%%%%%%%%%%%%%%%%%%%%%%%%

\begin{strip}
    \vspace*{-65pt}
    \centering
    \includegraphics[width=\textwidth]{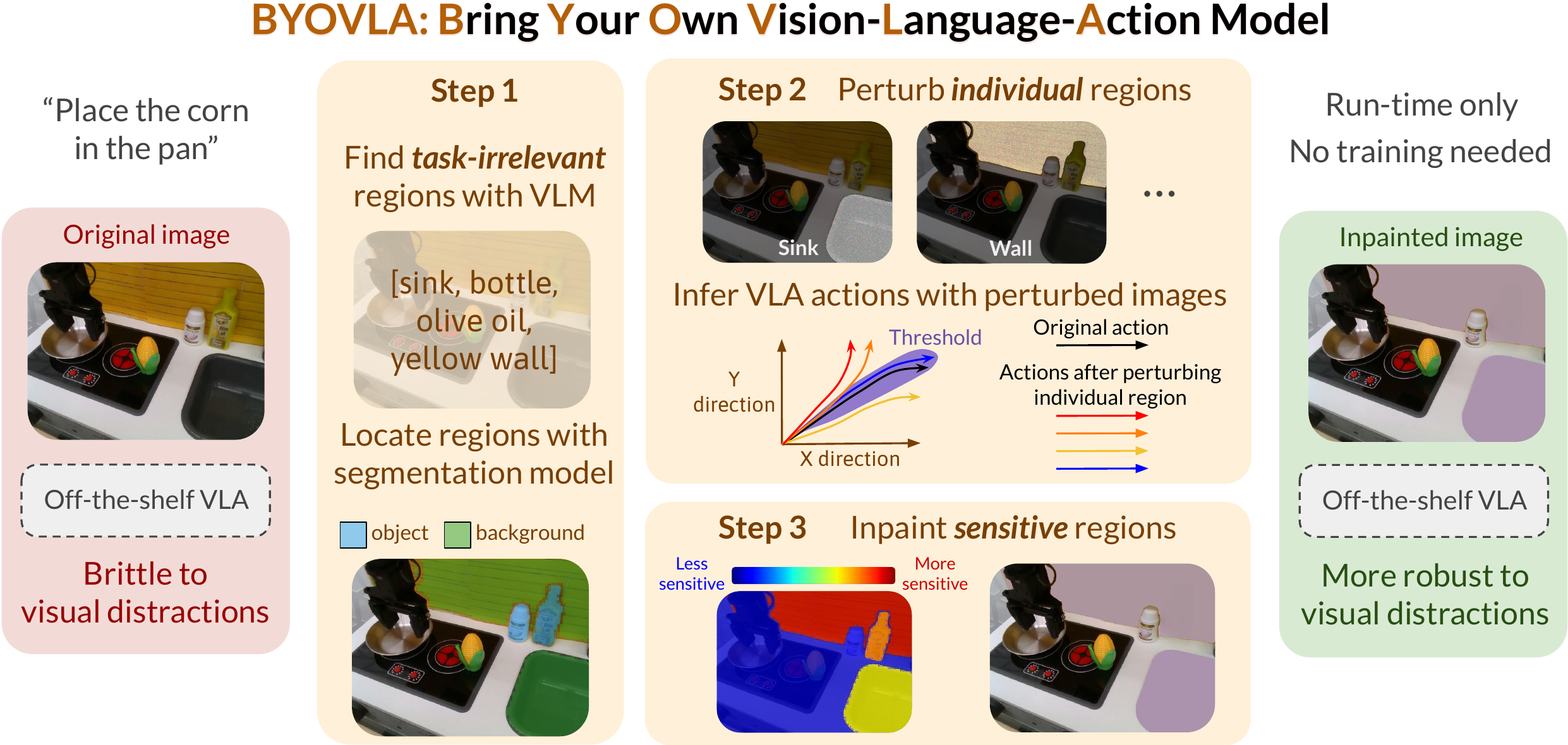}
    \captionof{figure}{We introduce \byovla{}: a simple and lightweight run-time intervention scheme for improving the performance of an arbitrary VLA model in the presence of task-irrelevant distractions. Our method identifies task-irrelevant regions in the visual observation and minimally modifies regions that the model is sensitive to in order to reduce sensitivity to distractors.}
    %\caption{We introduce \byovla{}: a simple and lightweight run-time intervention scheme for improving the performance of an arbitrary VLA model in the presence of task-irrelevant distractions. Our method identifies task-irrelevant regions in the visual observation and minimally modifies regions that the model is sensitive to in order to reduce sensitivity to distractors.}
    \label{fig:anchor}
\vspace{-5pt}
\end{strip}
%\twocolumn

\begin{abstract}
% \textcolor{blue}{0.5 Column}
Vision-language-action (VLA) models trained on large-scale internet data and robot demonstrations have the potential to serve as generalist robot policies. However, despite their large-scale training, VLAs are often brittle to task-irrelevant visual details such as distractor objects or background colors. We introduce \emph{Bring Your Own VLA} (\byovla): a run-time intervention scheme that (1) dynamically identifies regions of the input image that the model is sensitive to, and (2) minimally alters \emph{task-irrelevant} regions to reduce the model's sensitivity using automated image editing tools. Our approach is compatible with any off the shelf VLA without model fine-tuning or access to the model's weights. Hardware experiments on language-instructed manipulation tasks demonstrate that \byovla{} enables state-of-the-art VLA models to nearly retain their nominal performance in the presence of distractor objects and backgrounds, which otherwise degrade task success rates by up to 40\%. Website with additional information, videos, and code: \href{https://aasherh.github.io/byovla/}{https://aasherh.github.io/byovla/}.

\end{abstract}

%%%%%%%%%%%%%%%%%%%%%%%%%%%%%%%%%%%%%%%%%%%%%%%%%%%%%%%%%%%%%%%%%%%%%%%%%%%%%%%%
\section{Introduction}

% With the rise of large foundation models absorbing multiple data modalities such as language and image, 
A longstanding goal in robotics is to develop \emph{generalist} robot policies that can be instructed on the fly to perform tasks in diverse environments. Recently, vision-language-action (VLA) models trained with a combination of large-scale internet data and robot demonstrations have shown promise towards such generalization \cite{brohan2022rt, brohan2023rt, team2024octo, padalkar2023open}. These models leverage their internet-scale training to perform a broad range of visuomotor control tasks when prompted via natural language.
% Training large, transformer-based neural networks, with diverse and multimodal data, has led to the success of Vision-Language models (VLMs) as general-purpose visual reasoners. Recently, a similar recipe has been applied in training generalist robotic policies, known as Vision-Language-Action (VLA) models \cite{kim2024openvla, team2024octo, padalkar2023open}.  These models accept text instructions and camera observations as input and output actions for the robot to follow. Despite in their early stages, VLAs show promise in completing various tasks operating in manifold of initial conditions and environments. 

However, while existing VLAs show broad \emph{task} generalization, they fall short of their promise as generalist policies in terms of variations in \emph{environments}. Due to the complexity of real-world scenarios and the lack of robotic data at scale, state-of-the-art VLAs are brittle against marginal variations in the environments they were trained on. In particular, prior work \cite{brohan2022rt, brohan2023rt, team2024octo, kim2024openvla} and our experiments (Sec.~\ref{sec:experiments}) have shown a lack of \emph{visual} generalization; a small number of distractor objects or a mere change of background color, which leave the inherent task difficulty invariant, can \emph{drastically lower} the task success rates of VLAs.

While further scaling up data can potentially mitigate such performance drops, the effort required to collect such data and the computational resources required to fine-tune large VLAs (often with billions of parameters) is a strong deterrent. \emph{Can we design a lightweight and model-agnostic tool that does not alter the model weights, yet still improves robustness of VLAs to task-irrelevant objects and backgrounds?}

% In this work we are interested in improving the performance of an off-the-shelf VLA in the presence of task-irrelavent objects \emph{without additional training}. VLAs, with large number of parameters such as 7 billion \cite{kim2024openvla}, can be computationally expensive to train or fine-tune; thus we aim to design a lightweight, test-time-only tool that does not alter the model weights but improves the VLAs' robustness to visually distracting objects.

{\bf Contributions.} To this end, we propose \emph{Bring Your Own VLA} (\byovla): a run-time intervention scheme that improves visual generalization of off the shelf VLAs by minimally altering regions in the VLA's visual inputs in order to reduce sensitivity against visual distractors. The key idea is to identify (at run-time) which regions of the visual input the model is sensitive to using a \emph{visual sensitivity probe} that perturbs different segments of the visual input. \byovla{} queries a vision-language model (VLM) to identify which regions in the environment are task-irrelevant and alters a region using automated image editing tools (e.g., inpainting a distractor object) if the region is task-irrelevant \emph{and} the VLA is sensitive to it (Fig.~\ref{fig:anchor}).

\byovla{} can be applied to any VLA model without fine-tuning or access to the model's weights. Across multiple language-instructed manipulation tasks and varying distractor objects and backgrounds, \byovla{} improves task success rates by $20-40\%$ compared to the original VLA, while also significantly improving performance relative to baselines that perform run-time interventions (1) without accounting for the model's visual sensitivity or (2) assessing sensitivity via prior image attribution methods (e.g., GradCAM \cite{selvaraju2017grad}).

\section{Related Work}
\label{sec:related work}

\subsection{Vision-Language-Action (VLA) models} 
Building upon progress in foundation models for language and vision \cite{bommasani2021opportunities}, recent years have seen the rise of generalist vision-language-action (VLA) models \cite{brohan2022rt, brohan2023can, brohan2023rt, team2024octo, kim2024openvla} which show early promise in performing diverse tasks when prompted via natural language. This success has been enabled by a combination of existing internet data and large-scale efforts towards collecting human demonstration datasets such as Open X-Embodiment \cite{padalkar2023open} and DROID \cite{khazatsky2024droid}. 

Nevertheless, the complexity of real-world scenarios still overwhelms the amount of data available, and state-of-the-art generalist VLAs are often brittle against minor visual changes to the scene, such as the introduction of task-irrelevant objects or differing backgrounds. For instance, \cite{kim2024openvla} demonstrates that Octo \cite{team2024octo} --- a recently proposed VLA trained on Open X-Embodiment data --- has its task success rate dropped from 60\% to 29\% in visual generalization tasks consisting of object distractions and unseen object appearances or backgrounds.

\subsection{Improving policy robustness to visual distractors} There have been multiple lines of research for ameliorating the effect of visual distractors on a policy's performance. A straightforward and widely used method is to apply large-scale domain randomization to the visual observation \cite{tobin2017domain, laskin2020reinforcement}, often in the form of random noise or simple image manipulation such as cropping and translating. Recently, automated image editing tools (e.g., inpainting) have been used to generate diverse and more realistic backgrounds and object textures for data augmentation \cite{yuan2024learning, yu2023scaling, chen2024roviaug, mandi2022cacti}. These methods all apply such randomization \emph{during training}, whereas \byovla{} operates at run-time only and does not alter the model weights.

Another technique is to simply \emph{mask out} the background and possibly the irrelevant objects in the scene, either by learning a masking module end-to-end \cite{grooten2023madi} or using the segmentation object mask \cite{riedmiller2023less, stone2023open, zhu2023learning}. However, simple masking can make unrealistic observations which the model is subsequently trained on. A recent work \cite{yang2023transferring} uses a VLM to determine the relevant objects in the scene based on the task instruction, but again masks out \emph{task-irrelevant regions} and requires training with both the original and edited images.
\byovla{} does not require model re-training, and applies \emph{selective} masking based on model sensitivity. As current inpainting tools are imperfect, such minimal edits help mitigate artifacts potentially generated by the image editing process, keeping the transformed observations relatively realistic (as Fig.~\ref{fig:anchor} shows).

%Such minimal edits keep the edited observations relatively realistic and close to the training data (as \cref{fig:anchor} shows), thus not requiring additional training. Moreover, we mitigate artifacts potentially generated by the inpainting process, leading to the improved performance of \byovla{} compared to the baseline that inpaints all task-irrelevant objects in Sec.~\ref{sec:experiments}.

To our knowledge, \cite{miyashita2023roso} is the only other work that \emph{solely} performs run-time interventions of the camera observation for robot manipulation policies, but does not address visual distractions. Rather they focus on ensuring that \emph{task-relevant} objects are within the training distribution by inpainting novel ones with those seen during training. 

Other training strategies for improving model robustness include creating bottlenecks in the attention mechanism \cite{vaswani2017attention, bahdanau2014neural} of the policy architecture to train the policy to selectively focus on objects \cite{tang2020neuroevolution, james2022q}. Similar attention effects can also be achieved using information bottlenecks \cite{pacelli2020learning, igl2019generalization} or bisimulation-based state abstractions \cite{zhang2020learning, agarwal2021contrastive} for learning visual representations that only encode task-relevant information. Again these methods all require altering the training pipeline and are thus not compatible with VLAs off the shelf, unlike \byovla{}.

\subsection{Determining the task-relevant elements in the scene} As discussed above, previous work has investigated using VLMs \cite{yang2023transferring} or learning an end-to-end module \cite{grooten2023madi} to determine the task-irrelevant elements in the scene. \byovla{} also leverages the rich prior knowledge of VLMs to identify regions of the scene that are irrelevant, but visually manipulates them only if the model is sensitive to them. 

Our use of model sensitivity is also related to multiple \emph{attribution} methods --- usually used in image classification settings --- that seek to determine which part of the image (input) are most responsible for the model's output. Methods like SHAP \cite{lundberg2017unified} and LIME \cite{ribeiro2016should} determine how each input feature contributes to the output by learning a small model or a few parameters. Gradient-based methods such as GradCAM \cite{selvaraju2017grad} and SmoothGrad \cite{smilkov2017smoothgrad} compute how the model output changes as parts of the input observation are perturbed. However, these methods tend to be brittle and unreliable \cite{ghorbani2019interpretation, kindermans2019reliability}; specifically, the results can be sensitive to implementation details, such as the specific layer of the model network with respect to which the gradient is computed, or they may be entirely incorrect.
In contrast, the visual sensitivity probe we introduce in Sec.~\ref{sec:methodology} \emph{directly} measures changes in action outputs by perturbing different segments of the visual input. As our experiments in Sec.~\ref{sec:experiments} show, determining sensitivity using GradCAM \emph{does not} retain the base model's nominal performance in the presence of distractions. 

\begin{comment}
    \subsection{Information}
\begin{itemize}
    \item Foundation models in robotics enable vision-language-action (VLA) models, serving as generalist policies: Octo, OpenVLA, RT-1-X, RT-2-X, SayCAN, review articles (e.g., Ani's)
    \item For truly generalist policies, VLA models must be able to operate out-of-distribution (OOD). OOD detection with/in foundation models for robotics - Rohan's RSS work, robots that ask for help. \allen{I wonder if we should talk about OOD as the last (more tangential) point instead. VLA (or black-box policies in general) can be sensitive to distractors even if they are within the distribution.}
    \item There exist numerous attribution methods that seek to find which of a model's input/intermediate calculations were most salient in its final output
    \begin{itemize}
        \item Occlusion-perturbation-based: (SHAP, LIME) - model agnostic
        \item Gradient-based: (GradCAM, SmoothGrad) - model dependent
        \item Limitations - difficult to know whether correct, fragile, unreliable. As models get larger and more complex, its unclear whether a single attribution method will work. 
    \end{itemize}
\end{itemize}
\end{comment}

\section{Methodology}
\label{sec:methodology}

\subsection{Problem formulation}
Our goal is to improve the performance of a pre-trained VLA operating in environments with task-irrelevant visual distractions.
We consider policies $f(o_t, l)$ that take a language instruction $l$ as input in order to perform a visuomotor control task using RGB image observations $o_t$. In contrast to prior work (Sec.~\ref{sec:related work}), we propose a purely \emph{run-time} intervention that does not require any model fine-tuning or access to the model's weights. More formally, our goal is to process the raw observation $o_t$ to produce a new observation $\rho_f(o_t)$ that is then sent as input to the VLA to produce an action chunk (sequence):
\begin{equation}
    (a_t, \dots, a_{t+T_a}) \sim f(\rho_f(o_t), l),
\end{equation}
where $T_a$ is the action prediction horizon. 

The run-time intervention $\rho_f$ manipulates regions of $o_t$ that $f$ is sensitive to but that are \emph{irrelevant} to the task at hand, with the objective of recovering the nominal performance of the VLA \emph{in the absence} of visual distractors. We describe our pipeline for implementing $\rho_f$ in detail below.

\begin{algorithm}[t]
\caption{Bring Your Own VLA (\byovla)}\label{alg:byovla}
\begin{algorithmic}
\Require VLA model $f$, observation $o_t$, language instruction $l$, threshold $\tau$
%\State Initialize posterior distribution parameters $\mu, \Sigma \gets \mu_0, \Sigma_0$
    \State $\mathcal{R}_t \gets$ \Call{Task-Irrelevant Regions}{$o_t$}
    % \Comment{Use VLM \\ \quad \quad \quad \quad \quad \quad \quad \quad \quad \quad \quad \quad \quad \quad and segmentation model}
\State $(a_t, \dots, a_{t+T_a}) \gets f(o_t, l)$
\State Initialize $\rho_f(o_t) = o_t$
\For{each region $r \in \mathcal{R}_t$}
    \State $\tilde{o}_t \gets$ \Call{Perturb Region}{$o_t$, $r$}
    \State $(\tilde{a}_t, \dots, \tilde{a}_{t+T_a}) \sim f(\tilde{o}_t, l)$
    \State $\Delta_f(o_t, r) \gets$ Eq.~\eqref{eq:L2} \Comment{Calculate visual sensitivity} 
    \If{$\Delta_f(o_t, r) \geq \tau$}
        \State $\rho_f(o_t) \gets$ \Call{Image Editor}{$\rho_f(o_t)$, $r$}
    \EndIf
\EndFor
\State \Return $\rho_f(o_t)$
\begin{comment}
    \State Initialize $o_{\text{curr.}} = I$
\For{each mask  $m \in \mathcal{M}$}
\State Initialize $I_{\text{pert.}} = I$
\State $I_{\text{pert.}} \gets P(I_{\text{pert.}}, m)$
\State $ \delta_m \gets${$d\left( f(I), f(I_{\text{pert.}}) \right)$}
%\State $c(\mu,\Sigma) \gets KL(N(\mu,\Sigma)\Vert N(\mu_0, \Sigma_0)) - B(\alpha,\Hat{\alpha},\delta,D)$
\If{$\delta_m > \tau$} 
    \State $\rho_f(o_t) \gets V(\rho_f(o_t), m)$
\EndIf 
\EndFor
\end{comment}

\end{algorithmic}
\end{algorithm}

\subsection{Bring Your Own VLA}
Fig.~\ref{fig:anchor} and Algorithm~\ref{alg:byovla} provide an overview of our approach. Given a language instruction $l$ and an initial observation $o_0$, we first query a vision-language model (VLM) in order to identify visual regions that are \emph{irrelevant} to the task. At each time-step $t$ during policy execution, we then use a segmentation model to obtain corresponding masks for these irrelevant regions. A key component of our approach is to introduce a \emph{visual sensitivity probe} in order to identify which irrelevant segments the VLA $f$ is sensitive to. The final processed observation $\rho_f(o_t)$ is obtained by manipulating irrelevant regions (e.g., inpainting an object or changing the color of a background region) using automated image editing tools. We describe each of these components below. 
% Given a task, \byovla consists of three major steps: 1) find task-irrelevant regions in the image, 2) determine sensitive regions with visual sensitivity probe, and 3) transformation of the observation  (see Figure \ref{fig:anchor}). 
% \allen{label lines in algorithm and then refer to the lines when you introduce the steps below}

\textbf{Step 1: Localize task-irrelevant objects.} Semantic information about an image is readily captured by VLMs \cite{achiam2023gpt, Liu_2024_CVPR}, which we utilize to determine what regions in the initial image $o_0$ are task-irrelevant. We run the state-of-the-art GPT4-o model from OpenAI and prompt the model with few-shot exemplars (image observations paired with irrelevant regions in text). The output from GPT4-o is a string of region proposals in $o_0$ deemed task-irrelevant. The proposals are then provided to a segmentation model, 
% \cite{ren2024grounded, kirillov2023segany, minderer2022simple}
Grounded-SAM2 \cite{ravi2024sam2segmentimages, liu2023grounding, ren2024grounded, kirillov2023segany}, to localize and partition the regions at the pixel-level at every step of the rollout. Fig.~\ref{fig:anchor} depicts the outputs at each stage of the process. We consider static environments in our experiments, and thus GPT4-o is called once at initialization and the string of region proposals for the grounded segmentation model is held invariant during task execution. % The detections provided by the segmentation model, along with the corresponding masks, are utilized by the visual sensitivity probe. 

\textbf{Step 2: Apply visual sensitivity probe.} Given a set $\mathcal{R}_t$ of task-irrelevant regions from the VLM and segmentation model, we determine which of these impact the output of the VLA $f$. We quantify the sensitivity of $f$ to a region $r \in \mathcal{R}_t$ by perturbing the image in that segment to obtain $\tilde{o}_t$ and measuring the change in actions. Specifically, let $(a_t, \dots, a_{t+T_a}) \sim f(o_t, l)$ denote the predicted action chunk for the original observation $o_t$. Here, each action in the chunk corresponds to $(x, y, z, \phi, \theta, \psi, g)$: the relative displacements of the end-effector in translation and rotation, along with a gripper open/close state. Let $(\tilde{a}_t, \dots, \tilde{a}_{t+T_a})$ denote the action chunk for a perturbed observation $\tilde{o}_t$. After applying a single perturbation to region $r$ using a perturbation distribution described below, we sample $K$ action chunks $\{(\tilde{a}^k_t, \dots, \tilde{a}^k_{t+T_a})\}_{k=1}^K$. Additionally sampling $K$ action chunks $\{(a^k_t, \dots, a^k_{t+T_a})\}_{k=1}^K$ from the original image, we then compute an average weighted $L_2$-norm of the difference in actions $\Delta a^k_{t+t'} \coloneqq a^k_{t+t'} - \tilde{a}^k_{t+t'}$ (where $t' \in \{0,\dots,T_a\}$):

\vspace{-10pt}
\begin{equation}
    \Delta_f(o_t, r) \coloneqq \frac{1}{KT_a}\sum_{k=1}^K\sum_{t'=0}^{T_a}\sqrt{\langle w \Delta a_{t+t'}^k, \Delta a_{t+t'}^k \rangle},
    \label{eq:L2}
\end{equation}
where $w \in \mathbb{R}^7$ is a user-defined weighting vector. To perturb an image, we consider Gaussian blurring (smoothening) object distractions and adding Gaussian noise to background distractions for reasons explained below.

%While numerous methods exist to perturb an image, we consider Gaussian blurring (smoothening) for object distractions and Gaussian noising for background distractions. Our choice of perturbation reflects offline evaluations of Octo in environments from the BridgeV2 dataset \cite{walke2023bridgedata}, which includes the physical kitchenette we consider in our experiments. 
% Empirically, we also find that this simple choice of perturbation distribution is sufficient to capture model sensitivity. 

\textbf{Determining the sensitivity threshold.}
If the quantity $\Delta_f(o_t, r)$ in Eq.~\eqref{eq:L2} is greater than a threshold $\tau$ for a region $r$ from the segmentation model, we intervene on that region.
To determine $\tau$ for object distractions, we utilize the first observation from 60 environments in BridgeV2 and apply Gaussian blurring to the task-irrelevant object regions in the image following Step 1 above. Computing Eq.~\eqref{eq:L2} with $w$ as the indicator function for translational components, and then taking the third quartile, we arrive at a value of approximately $0.005\si{\meter}$. Since different environments in BridgeV2 are of different physical scales, we adjust the threshold by rolling out a few trials in our kitchenette, arriving at a threshold value for object distractors of $\tau=0.002\si{\meter}$. 

For background distractions, we repeat the same procedure and arrive at a threshold of $\tau=0.001\si{\meter}$. We add Gaussian noise to the the RGB channels of the observation, instead of Gaussian blurring, as we find blurring too weak to cause a substantial deviation in trajectories for these regions. The values for $\tau$ described above are used for all experiments.  
% Additional numerical details are described in Section \ref{sec:experiments}. 

%A model is expected to be sensitive to regions in an image pertaining to the task or goal; conversely, if a model is sensitive to regions unrelated to the task, it suggests the model is attending to spurious states. The premise of this step of BYOVLA, coined "visual sensitivity probing," is to find regions of ultra-sensitivity and intervene such that they are no longer. 

%Perturbations to a region in $I$, defined by mask $m$, can be any pixel-level manipulation. In our work, we consider adding Gaussian noise and Gaussian blurring to a region, but other methods would suffice so long as the induced perturbation leads to model sensitivity. 

\textbf{Step 3: Transform the image.} %If Eq.~\eqref{eq:L2} is greater than the threshold $\tau$ for a region defined by mask $m$ from the segmentation model, the observation is intervened upon. The specific image transformation is dependent upon whether the region is classified as an object or background distraction.
The specific image transformation is dependent upon whether the region is classified as an object or background distraction, which can be determined by the VLM. If the region is an object distraction, a vision model capable of inpainting is called to remove it from the image; in our experiments, we use Inpaint Anything~\cite{yu2023inpaint}. 

If the region is a background distraction, the RGB pixels in that region are simply altered such that $\Delta_f(o_t, r) < \tau$. Recall the intuition that we would like the transformed observation to better match the training data. Since the distribution of colors seen during training is hard to specify a priori, we choose a random, neutral color to inpaint the background region with, recalculate Eq.~\eqref{eq:L2} for the inpainted image and inpainted-plus-noised image, and repeat until $\Delta_f(o_t, r)$ is below the threshold. 
% We find this simple recipe works well in practice. 

\begin{comment}
\subsection{Information}
\begin{itemize}
    \item Intuition/hypothesis: a model's input-output sensitivity to task-irrelevant features is highly correlative with poor performance. For instance, this includes task-irrelevant distractor objects, novel background colors, etc. 
    \item Approach: we will consider VLA models which accept a task string and input image as input, and produce action commands.
    \begin{itemize}
        \item use LLM's semantic knowledge to determine task irrelevant features of an input image
        \item use powerful VLMs to partition input image
        \item determine model's sensitivity to original and "perturbed" version of image 
        \item threshold based upon environmental details, e.g., size/expert knowledge
        \item remove or intervene on regions which the model is too sensitive to
    \end{itemize}
\end{itemize}

\subsection{Figures}
\begin{itemize}
    \item Flowchart. Maybe have the same as in Intro?
\end{itemize}
\end{comment}

\section{Experiments}
\label{sec:experiments}
We evaluate \byovla{} with two state-of-the-art open-source VLA models: Octo-Base \cite{team2024octo} (93M parameter transformer-based diffusion policy) and OpenVLA \cite{kim2024openvla} (7B parameter transformer-based autoregressive policy). The tasks considered are "put the carrot on yellow plate" and "put the eggplant in the pot," which take place in a toy kitchen environment from the BridgeData V2 dataset \cite{walke2023bridgedata} and are the representative tasks used for evaluation in \cite{team2024octo} and \cite{kim2024openvla}. 

\begin{figure}[h]
    \centering
    \includegraphics[width=\columnwidth]{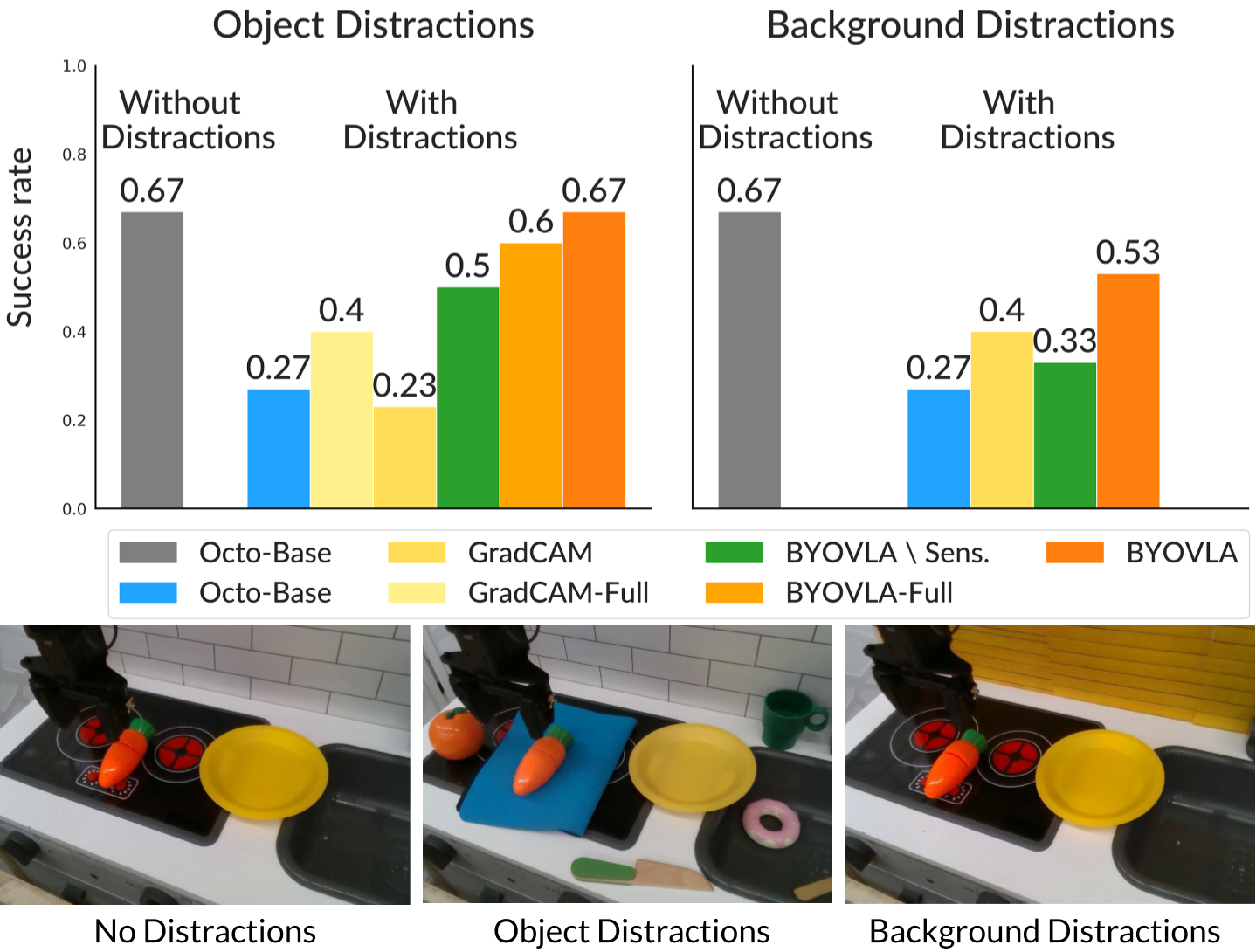}
    \caption{First row: task success rates for \byovla{} with Octo on language instruction "place the carrot on yellow plate." Second row: kitchenette environment from BridgeV2 dataset with and without object and background distractions.}
    \label{fig:octo_object_background}
\end{figure}

{\bf Environments and hardware setup.} In our experiments, we consider object and background environmental distractions. \textbf{Object distractions} include items commonly found in a kitchen environment but which are irrelevant for task completion. Importantly, object distractions do not affect the trajectory required by the robot to reach the goal state. In each task, 5-7 object distractions are added to the domain; these objects are selected from the BridgeData V2 catalog of objects and are thus \emph{not adversarial} in nature. \textbf{Background distractions} include changes to the appearance of the scene background that are irrelevant to the task. Fig.~\ref{fig:octo_object_background} depicts object and background distractions in our kitchen environment for the task "put the carrot on yellow plate": addition of an "orange fruit" is an object distraction whereas changing the tiling color to yellow is a background distraction. In general, distractions are chosen to be realistic while weakening VLA performance in order to assess the benefits of \byovla.  

Following the hardware experiments from Octo \cite{team2024octo} and OpenVLA \cite{kim2024openvla}, all policies are evaluated on the Widow X 250S robotic arm in accordance with the setup prescribed by \cite{walke2023bridgedata}. Camera angles and object/background distractions are held constant during all trials. A threshold of 0.7 for the gripper state is set to determine when to open or close the end-effector. The Widow X is initialized at the same position above the task object; consequently, $w$ is kept as the indicator function for translational components in Eq. ~\eqref{eq:L2} for all experiments, the rationale being that we find rotational and gripper commands insignificant to task success if the robot starts from this position. In accordance with \cite{team2024octo} and \cite{kim2024openvla}, the task object's initial position is varied 1-3cm from a central position between trials. Unless otherwise stated, 15 trials for all baselines are completed. 

\textbf{Run-time.} In general, the overhead incurred by \byovla{} is reliant on (1) the inference speed of the underlying foundation models and (2) the number of task-irrelevant regions to manipulate. Queries to the VLM (GPT4-o) for determining task-irrelevant objects on average take less than three seconds to complete and cost less than one cent with five few-shot exemplars; we assume static environments, so this query is executed once at the beginning of the episode. Irrespective of inpainting, Octo-Base and OpenVLA run at 13Hz and 6Hz, respectively, on a NVIDIA GeForce RTX 4090 GPU \cite{team2024octo, kim2024openvla}, which we used in our experiments. In our (object distraction) experiments with Octo, Steps 1-3 above, without calling GPT4-o, take roughly 2 seconds to complete. All time measurements were averaged over 15 trials. 

%We found that inpainting a single object with Grounded-SAM2 took less than a second. All time measurements were averaged over 15 trials run on a NVIDIA GeForce RTX 4090 GPU.

\begin{figure*}
    \centering
    \includegraphics[width=\textwidth]{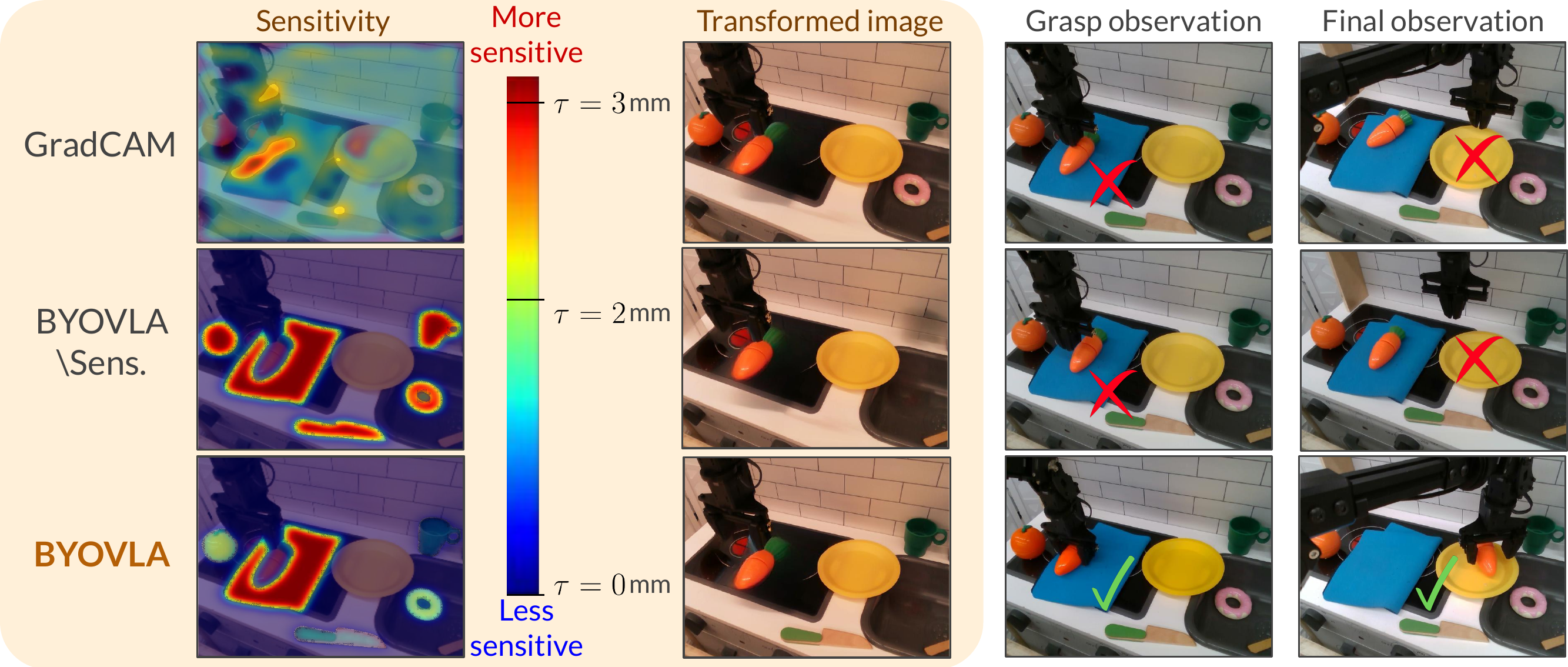}
    \caption{
    % First column: heatmaps depicting which task-irrelevant regions each methodology claims the VLA is sensitive to. Second column: inpainted image regions with sensitivity threshold $\tau$. Remaining columns: observations of the task object grasp and final state. The GradCAM and \byovla{}$\setminus$Sens. baselines claim the VLA is sensitive to regions different from what \byovla{}'s visual sensitivity probe predicts, e.g., the green wooden knife.
    First column: heatmaps showing the regions each method deems the VLA is sensitive to. Second column: inpainted image regions with sensitivity threshold $\tau$. 
    % Remaining columns: observations of the task object grasp and final state. 
    % GradCAM and \byovla{}$\setminus$Sens. considers sensitive regions different from those from \byovla{}, e.g., the green wooden knife.
    \byovla{} inpaints the the blue towel, orange, and donut, and then successfully grasps the carrot and puts it on the plate (last two columns), while \byovla{}$\setminus$Sens. additionally inpaints the green knife and cup, but fails the task. GradCAM fails to capture the model sensitivity to most irrelevant objects and thus also fails.
    }
    \label{fig:octo_gradCAM}
\vspace{-10pt}
\end{figure*}

%\byovla{}$\setminus$

\textbf{Baselines.} We evaluate \byovla{} against these baselines: (1) the original VLA policy; (2) \byovla{} without the visual sensitivity probe --- labeled in Figs.~\ref{fig:octo_object_background} -- \ref{fig:openVLA} as \byovla{}$\setminus$Sens. --- where we manipulate \emph{all} regions of the image deemed task-irrelevant by the VLM. 
For our experiments with Octo, we consider (3) a GradCAM-based \cite{selvaraju2017grad} baseline, where we replace our visual sensitivity probe with GradCAM to attribute what regions in the image are most important for model output and manipulate those regions if they are deemed task-irrelevant by the VLM. In particular, we calculate the cross-attention values and gradients between the image and task tokens, which we then average across the attention heads and task tokens. We perform this operation halfway through the overall transformer architecture (layer 6 in Octo-Base), which is motivated by recent work in mechanistic interpretability suggesting that intermediate layers of transformer-based architectures contain salient features \cite{templeton2024scaling, gao2024scaling, elhage2022solu, chefer2021transformer}. To determine which image regions to manipulate, we compute the difference between maximal and minimal GradCAM scores and retain the pixel locations corresponding to the top quarter fraction. See Fig.~\ref{fig:octo_gradCAM} for an example of the regions deemed salient. 

We few-shot prompt GPT4-o with five in-context examples including an image containing our kitchenette environment with distractions, along with a list of the task-irrelevant regions in the image. At run-time, we only query GPT4-o with the initial observation $o_t$ and ask for the task-irrelevant regions. In principle, the distinction between an object and background distraction can be determined by GPT4-o. However, in our experiments, we consider the effect of object and background distractions separately, and hence do not prompt GPT4-o to distinguish between them. Unless stated otherwise, determination of task-irrelevant regions with GPT4-o and visual sensitivity probing is done once at initialization and fixed throughout the episode. While one can easily perform these operations at every step, our experiments for the tasks considered in this work demonstrate no \emph{additional} benefit (see Fig. \ref{fig:octo_object_background}). We discuss this further in Sec. \ref{sec:conclusions}.

\subsection{Evaluation with Octo-Base}

{\bf Task and distractors.} \byovla{} is first evaluated with Octo-Base on the task "place the carrot on yellow plate." The bottom images of Fig.~\ref{fig:octo_object_background} depict the environmental distractions present in the kitchenette environment. The left scene depicts the environment without distractions, and the middle scene showcases five task-irrelevant objects: orange, blue towel, knife, green cup, and donut. The rightmost scene demonstrates the yellow tiling background distraction mentioned previously.

{\bf Implementation details.} For the object distractor experiments, 30 trials are completed for each baseline. For visual sensitivity probing, we Gaussian blur object regions with a kernel size of 25, and add Gaussian noise $\eta \sim \mathcal{N}(0, \sqrt{0.075})$ to the RGB channels for background regions. The threshold is set to $\tau=2$mm for objects and $\tau=1$mm for background regions for reasons described in Sec. \ref{sec:methodology}.  We transform the input image with a warm filter in order to better match our physical operating conditions to Octo's training environments. We utilize the full extent of Octo's action chunking capability ($T_a=3$), which we find most effective for achieving the baseline success rate reported in \cite[Appendix]{team2024octo}. In calculating Eq.~\eqref{eq:L2}, $K=5$ rollouts are sampled. 

\begin{comment}
    \begin{SCfigure*}
    \centering
    % \includegraphics[width=\columnwidth, scale=0.8]
\includegraphics[width=0.9\textwidth]{Figures/gradCAM_figure.png}
    % \caption{Left column: comparison of GradCAM, \byovla{}$\setminus$Sens., and \byovla{} sensitivity baselines from Octo-Base object distraction experiments. Middle-left column: transformed image based upon which objects were deemed sensitive. Middle-right column: observations at a later time step showcasing an unsuccessful early grasp for GradCAM and missed target for \byovla{}$\setminus$Sens., respectively. However, a successful grasp for \byovla{} is seen. Right column: final observations of the trial.} 
    \caption{Four columns: (1) sensitivity (2) ... (3) ... (4). We find...}
    \label{fig:octo_gradCAM}
\end{SCfigure*}
\end{comment}

%According to GradCAM, the VLA is only attending to task relevant regions; yet, the policy's brittleness to object distractors implies the model is sensitive to them.  \byovla{}$\setminus$Sens., 

 %First row: GradCAM output from Octo-Base model evaluated at the cross-attention mechanism between language tokens and image tokens at layer 6, along with its transformed image $\rho_f(o_{t=1})$ and resultant trajectory at $t=6$. According to GradCAM, the VLA is only attending to task relevant regions; yet, the policy's brittleness to distractors implies the model is attending to these. Second row: sensitivity heat map from \byovla, along with its transformed image $\rho_f(o_{t=1})$ and resultant trajectory at $t=6$. \byovla{} yields a successful grasp and task completion.}

{\bf Results.} As Fig. \ref{fig:octo_object_background} shows, task-irrelevant object distractions drop Octo-Base's success rate by 40\%. Manipulating the image via inpainting following GradCAM or \byovla{}$\setminus$Sens fails to retain the nominal task success rate of 67\% without distractions present, whereas \byovla{} is able to.

With object distractions, we investigate whether applying the visual sensitivity probe at \emph{every} step with subsequent image manipulation, denoted by "-Full", offers any additional benefit over determining sensitivity at initialization \emph{only}: Fig.~\ref{fig:octo_object_background} suggests not, which we hypothesize is due to the static environments and relatively simple tasks considered. While we anticipate updating model sensitivity will offer benefit in dynamic environments, unless otherwise stated, we only run the visual sensitivity probe at the initial time-step. 

For background distractions, a similar trend in performance is observed. In this case, the GradCAM baseline slightly outperforms Octo-Base and \byovla{} achieves the best performance, raising Octo's task success rate by $\sim$25\%.

In general, the most common failure modes observed across all distractions and all baselines, \byovla{} included, are early grasping of the task object and missing the task object upon approach (see Fig.~\ref{fig:octo_gradCAM}). These failure modes are especially reflective of the effect of distractions since we only vary the initial position of the task object by 1-3cm from its central location between trials. Since this variation is smaller than the gripper's width when open, it highlights the deleterious effect distractions have on the policy. 

The improvement of \byovla{} over \byovla{}$\setminus$Sens. is likely attributable to the presence of distractors in the training data, e.g., the BridgeV2 dataset, which forms a significant portion of Octo's training data. By removing all such distractions from the input image, the distribution shift induced by inpainting is likely responsible for the policy's failure. On the other hand, we find that GradCAM is not attending to all regions relevant for Octo's output; otherwise, the success rate would have approached the nominal. The trial in Fig.~\ref{fig:octo_gradCAM} depicts results in a failed trajectory due to an early grasp of the object when inpainting what GradCAM says Octo is sensitive to, whereas running \byovla{} results in a successful trajectory.

% [width=\textwidth]
% [width=\columnwidth]

\subsection{Evaluation with OpenVLA}

\begin{figure}
    \centering
    \includegraphics[width=\columnwidth]{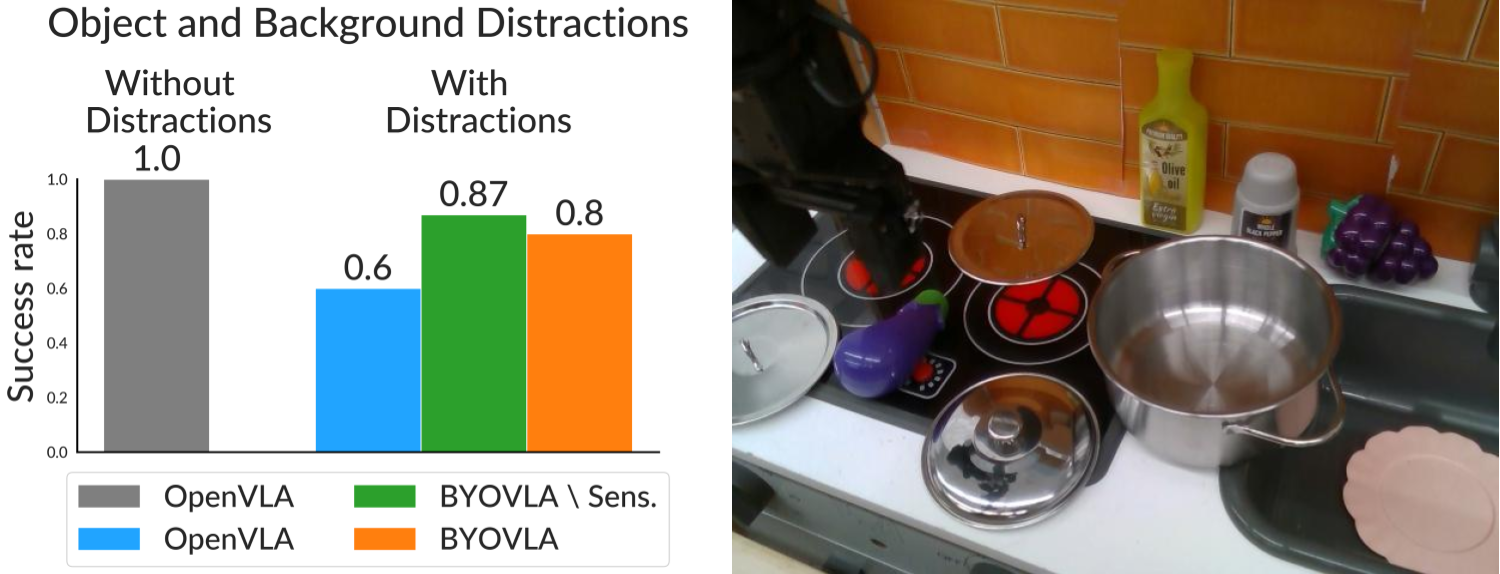}
    \caption{First column: task success rates for \byovla{} with OpenVLA on language instruction "put the eggplant in the pot." Second column: kitchenette environment from BridgeV2 dataset with distractions.}
    \label{fig:openVLA}
\vspace{-15pt}
\end{figure}

{\bf Task and distractors.} We next study \byovla{} with OpenVLA on the task "put the eggplant in the pot," where we seek to answer two questions: (1) to what extent is \byovla{} model-agnostic, and (2) can \byovla{} offer any benefit to policies that build on vision-language models pretrained on large-scale internet data? We again utilize distractor objects from the BridgeV2 dataset, which accounts for roughly a sixth of total data that OpenVLA was trained on \cite{kim2024openvla}. The rightmost image in Fig. \ref{fig:openVLA} depicts the task-irrelevant objects: three silver lids, olive-oil, black pepper, grapes, and a pink plate. Brown bricks are chosen to contrast the original white tiling as a background distraction.  %OpenVLA exhibits significantly greater visual generalization capabilities than Octo \cite{kim2024openvla}, which will further elucidate any benefits offered by BYOVLA as models scale in size and complexity. 

{\bf Implementation details.} Object (background) regions are again Gaussian blurred (Gaussian noised) for visual sensitivity probing with a threshold of $\tau=2$mm for objects and $\tau=1$mm for background regions (identical to the Octo experiments). Unlike Octo, OpenVLA does not action-chunk its commands. Moreover, unlike Octo that uses a diffusion policy and outputs stochastic action chunks, OpenVLA deterministically outputs the autoregressed action; therefore, only one rollout ($K=1$) is used for evaluating Eq.~\eqref{eq:L2}.

{\bf Results.} As Fig.~\ref{fig:openVLA} shows, while OpenVLA nominally achieves a perfect task success rate, the presence of distractions dropped its performance by 40\%. \byovla{} and its variant \byovla{}$\setminus$Sens. both improve the baseline performance by 20-25\% in the presence of distractions. The most common failure mode observed for OpenVLA across all baselines, \byovla{} included, is the tendency to not command any changes in position after successfully grasping the task object, e.g., not lifting the eggplant off the stove.

While \byovla{} does not retain the nominal success rate of OpenVLA with distractions, the environment depicted in Fig.~\ref{fig:openVLA} is significantly more cluttered than the environment for Octo's experiments. We find that even with few-shot prompting, GPT4-o often fails to locate all task-irrelevant objects, namely the lids located on the stovetop; this likely contributes to \byovla{} not reattaining OpenVLA's nominal performance.

% \begin{table}[h]
% \small
% \begin{center}
% \begin{tabular}{ccc}
%     \toprule
%     Notation & Description & Range\\
%     \midrule
%     $\mu$ &... & $[0.25,0.4]$\\
%     $h$ & ... & $[1.5\text{cm}, 2.5\text{cm}]$\\
%     \bottomrule
% \end{tabular}
% \caption{...}
% \label{tab:fetch-results}
% % \vspace{-10pt}
% \end{center}
% \end{table}
% \TODO{make a table for success rates and stopping criterion}

\begin{comment}
    \section{Discussion [0.75 Column]}
Key Findings: 
\begin{itemize}
    \item Sensitivity analysis for object distractors improves over naive outpainting everything deemed task-irrelevant. This could be because 1) current open-source models don't always produce perfect results and 2) black-box nature of VLA models, often trained with distractors present, means that eliminating all distractors could take the input further OOD than otherwise 
    \item 
\end{itemize}
Pros and Cons of approach
\begin{itemize}
    \item Pro: model-agnostic - can be deployed readily on any VLA model without access to model internals.
    \item Pro: can improve performance of model without needing to resort to larger, more expensive models, e.g., OpenVLA
    \item Con: Success of LLM and VLM pipeline is often mediocre, and when successful, not perfect
    \item Con: Threshold is a crucial hyperparameter that will likely need redetermined between realizations.
\end{itemize}

\end{comment}

\section{Conclusion and discussions}
\label{sec:conclusions}

We present \byovla{}: a run-time intervention scheme that dynamically determines task-irrelevant regions that an arbitrary VLA is sensitive to and minimally alters the image with automated image editing tools to improve policy performance in the presence of object and background distractions. \byovla{} is applicable off the shelf and does not require access to the VLA's weights. Experiments show that \byovla{} allows VLAs to nearly retain their nominal performance in the presence of task-irrelevant distractors, which otherwise drop the task success rate by up to 40\%.

\textbf{Limitations and future work:} The success of \byovla{} is reliant upon orchestrating different foundation models into a common pipeline, which presents challenges with integration. One limitation of our approach is the distinction between object and background distractions; while VLMs like GPT4-o can in principle discern between the two, our experiments primarily focus on cases where objects and backgrounds are separated to maximize the performance of GPT4-o in determining task-irrelevant regions. We expect that as VLMs become more capable, this aspect of \byovla{} improves.

Moreover, the regions proposed by the VLM are not guaranteed to be found with a separately trained segmentation model, and the choice of threshold $\tau$ is a hyperparameter of our method that requires a few real-world deployments to fine-tune for best results. Future work will consider how to better choose a threshold for a given environment, e.g., using conformal prediction \cite{angelopoulos2021gentle, shafer2008tutorial} to bound the false positive rate of detecting sensitive regions. In addition, we plan to explore more sophisticated inpainting schemes for background regions that seek to replace the background at deployment time with backgrounds from the VLA's training data.
% leverage the distribution of backgrounds from the Open X-Embodiment dataset. 
Finally, we only consider static environments in this work and plan to apply \byovla{} in \textit{dynamic} environments, e.g., with a mobile robot or human behavior in the scene, where task-relevancy may change during policy execution. Therein, we expect that running the entire pipeline of \byovla{} at every time-step will offer advantages not capitalized on here. 

%Our experiments focus on common household objects likely present in the training data of both the VLM and segmentation model, so this issue does not affect our results. However, in more complex environments, the probability that a segmentation model locates all of the VLM's region proposals may decrease. This issue may be alleviated with the future development of VLMs that are directly capable of segmentation. 

\vspace{5pt}

Overall, we believe that \emph{run-time interventions} --- as a form of test-time compute --- represent an underexplored avenue for significantly improving the base capabilities of VLAs without additional training, and we hope that the work presented here spurs further research in this area.

% \addtolength{\textheight}{-12cm}   % This command serves to balance the column lengths on the last page of the document manually. It shortens the textheight of the last page by a suitable amount.This command does not take effect until the next page so it should come on the page before the last. Make sure that you do not shorten the textheight too much.

%%%%%%%%%%%%%%%%%%%%%%%%%%%%%%%%%%%%%%%%%%%%%%%%%%%%%%%%%%%%%%%%%%%%%%%%%%%%%%%%

% \section*{ACKNOWLEDGMENT}

% The

%%%%%%%%%%%%%%%%%%%%%%%%%%%%%%%%%%%%%%%%%%%%%%%%%%%%%%%%%%%%%%%%%%%%%%%%%%%%%%%%

% References are important to the reader; therefore, each citation must be complete and correct. If at all possible, references should be commonly available publications.

\newpage
\bibliography{root}
\bibliographystyle{IEEEtran}

%%%%%%%%%%%%%%%%%%%%%%%%%%%%%%%%%%%%%%%%%%%%%%%%%%%%%%%%%%%%%%%%%%%%%%%%%%%%%%%%

\newpage

%%%%%%%%%%%%%%%%%%%%%%%%%%%%%%%%%%%%%%%%%%%%%%%%%%%%%%%%%%%%%%%%%%%%%%%%%%%%%%%%
\section*{Appendix}
\subsection{VLM Prompting}
We provide an example of few-shot prompting GPT4-o to determine the task-irrelevant objects from the experiment "place the carrot on yellow plate" in Sec. \ref{sec:experiments}. Each demonstration consists of a list of task-irrelevant objects, backgrounds, a natural language instruction, and an initial image. We utilized the same toy kitchen environment for all examples, but arbitrary environments from any offline dataset should suffice, e.g., BridgeV2 \cite{walke2023bridgedata}.

\begin{lstlisting}[breaklines,  escapeinside=||, label=listing:vlm-prompt]
You are an assistant helping a robot determine what objects in the image are relevant for completing its task. You will be shown some text and images. 

Example 1. Task: 'put strawberry in the bowl'

["hot dog", "broccoli"]
["white counter", "wall", "stovetop", "black sink"]

|\includegraphics[width=\columnwidth]{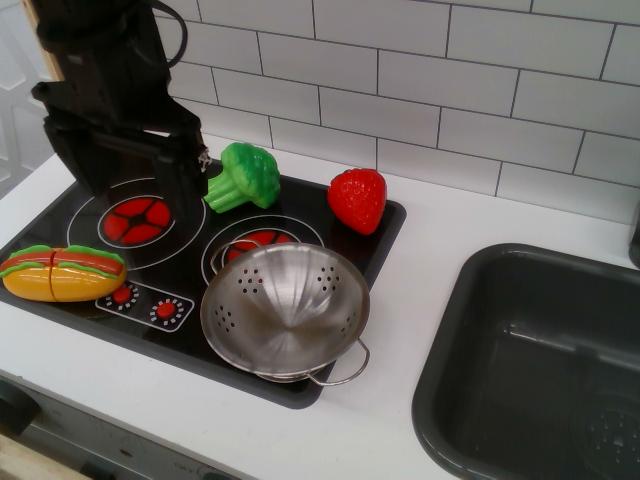}|

Example 2. 
|$\vdots$|

Example 5. Task: 'pick up the tomato'

["pizza", "olive oil", "lid"]
["wall", "counter", "stove", "sink"]

|\includegraphics[width=\columnwidth]{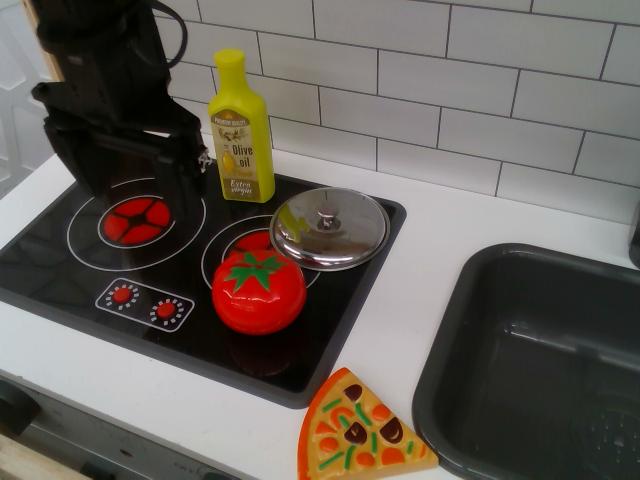}|

The robotic arm in the image is given the following task: 'place the carrot on yellow plate.'
|\includegraphics[width=\columnwidth]{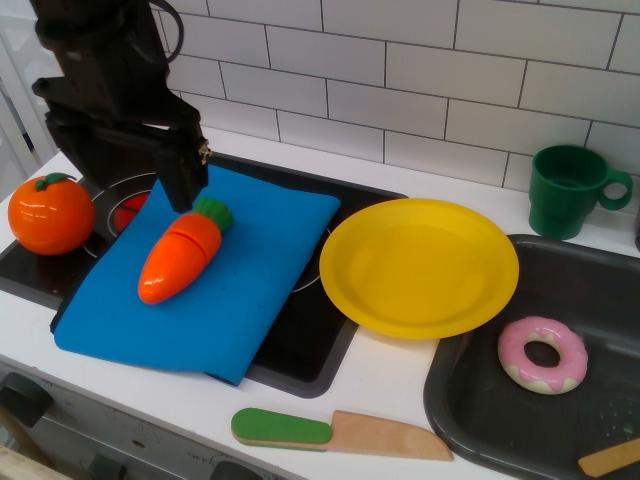}|
Provide a list of objects in the image that are not relevant for completing the task, called 'not_relevant_objects'. Then provide a list of backgrounds in the image that are not relevant for completing the task, called 'not_relevant_backgrounds'. Give your answer in the form of two different lists with one or two words per object.

\end{lstlisting}

\subsection{GradCAM Calculation}
GradCAM is a model-specific calculation and we study its efficacy for sensitivity probing with Octo-Base as our VLA. 

Given observation $o_t$, we perform one forward-pass through Octo-Base to obtain the multi-headed, cross attention weights $A^{(h,l)}_{i,j}$ between the task tokens $i$ and image tokens $j$ at head $h$: any extraneous tokens between 'wrist' or 'readout' tokens are removed (see \cite{team2024octo} for further discussion regarding these token types). We then select the attention weights corresponding to the specific transformer layer $l$ of interest and average across the task tokens dimension, yielding $A^{(h)}_{j}$. 

For the gradient calculation, we perform one backward-pass through the VLA with the current observation $o_t$ to obtain the multi-headed, cross attention gradient weights $\partial A^{(h,l)}_{i,j}$ between the task and image tokens, again removing extraneous tokens. We then average the gradients across the task token dimension at layer $l$ and resize into a square array of dimension $\sqrt{d_{o_t}}$, the pixel dimensions of the current observation (256 in the case of Octo-Base). This yields the gradient calculation $\partial A^{(h)}_{j}$. 

Finally, we compute the standard GradCAM calculation from \cite{selvaraju2017grad} and average across the $H$ attention heads $h$, yielding

\begin{equation}
    L_{\text{GradCAM}, j} = \frac{1}{H} \sum_{h} \left (\partial A^{(h)}_{j} \right) A^{(h)}_{j}.
    \label{eq:GradCAM}
\end{equation}

In Octo-Base, there are 256 image tokens $j$, so the resultant GradCAM calculation in Eq. ~\eqref{eq:GradCAM} is a 16x16 square matrix.

\begin{figure}[ht]
    \centering
    \includegraphics[width=\columnwidth]{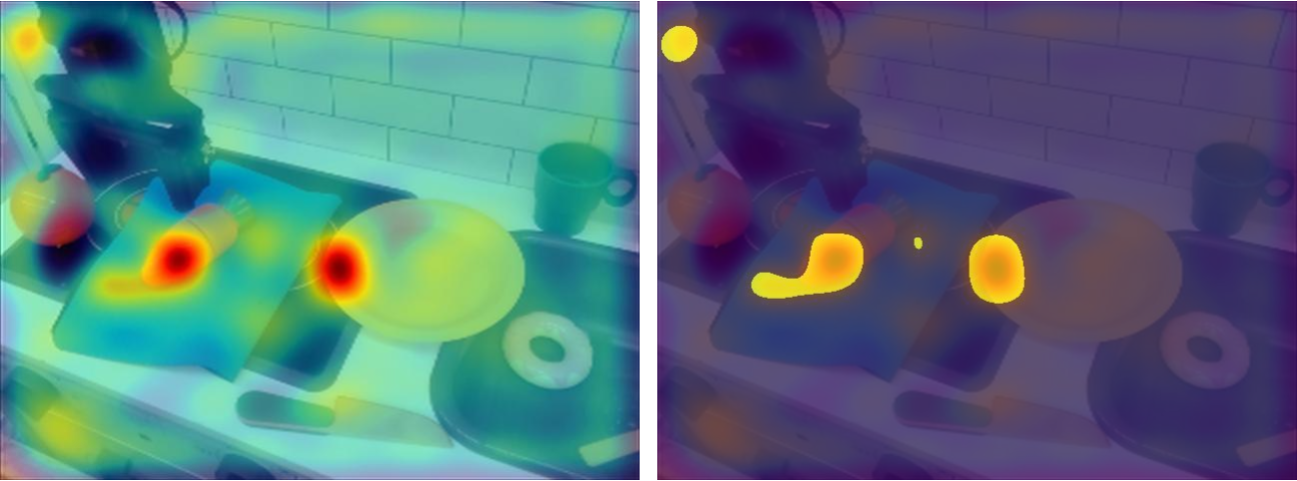}
    \caption{Left: GradCAM output at layer 6 with Octo on language instruction "place the carrot on yellow plate." Right: mask used to select what objects were attended to by keeping the top quarter of GradCAM values.}
    \label{fig:app_gradcam_mask}
\end{figure}

We apply a Gaussian blur to the GradCAM mapping to create a smoother image. An example of the GradCAM output and mask with thresholding a quarter of the greatest values can be seen in Fig. \ref{fig:app_gradcam_mask}

\subsection{Foundation Model Hyperparameters}

\subsubsection{Selection of Threshold $\tau$}

\begin{figure}[ht]
    \centering
    \includegraphics[width=\columnwidth]{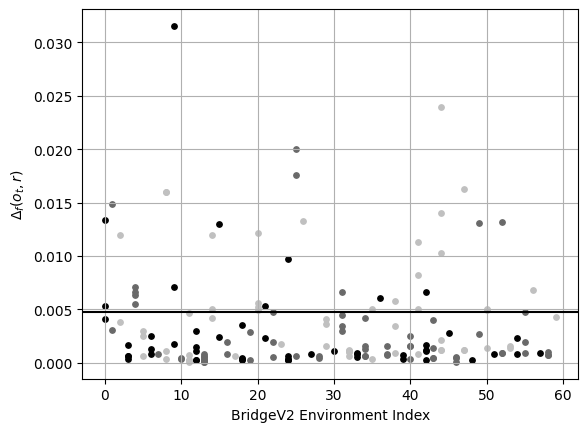}
    \caption{Computation of Eq.~\eqref{eq:L2} for BridgeV2 environments using the methodology prescribed in Sec. \ref{sec:methodology}. Each datapoint for a particular environment corresponds to a single perturbed task-irrelevant object. The horizontal line corresponds to the third quartile, arriving at a value of approximately $\tau = 0.005$.}
    \label{fig:tau_objects}
\end{figure}

Selection of $\tau$ is described in Sec. \ref{sec:methodology}, which is based upon applying \byovla{} offline to environments from the BridgeV2 dateset. Fig. \ref{fig:tau_objects} showcases the third quartile, giving us an initial threshold of 5 [mm].

\subsubsection{Segmentation Model}

\begin{figure}[h]
    \centering
    \includegraphics[width=\columnwidth]{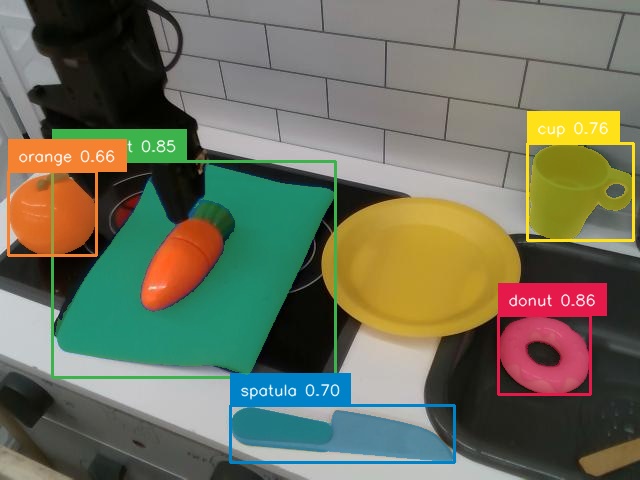}
    \caption{Output from Grounded-SAM2 with Octo on language instruction "place the carrot on yellow plate." The output from the VLM was the list ['orange', 'blue mat', 'spatula', 'donut', 'cup'].}
    \label{fig:app_GS2_output}
\end{figure}

We utilized Grounded-SAM2 with a confidence score of 0.4 for the box and text thresholds. See Fig. \ref{fig:app_GS2_output} for a demonstration of the output. For removing objects in image with \cite{yu2023inpaint}, we used a mask dilation size of 10 which was empirically chosen to minimize image aberrations.  

\subsection{Ablations}
\subsubsection{Sampling K Observations}
Rather than sampling K action chunks from the VLA for a single image, we investigated sampling K observations and a single action chunk. For object distractions, this corresponded to randomly sampling K kernel sizes between [15,30]. For background distractions, this corresponded to creating K observations with Gaussian noise, which is intrinsically stochastic.

\begin{figure}[h]
    \centering
    \includegraphics[width=\columnwidth]{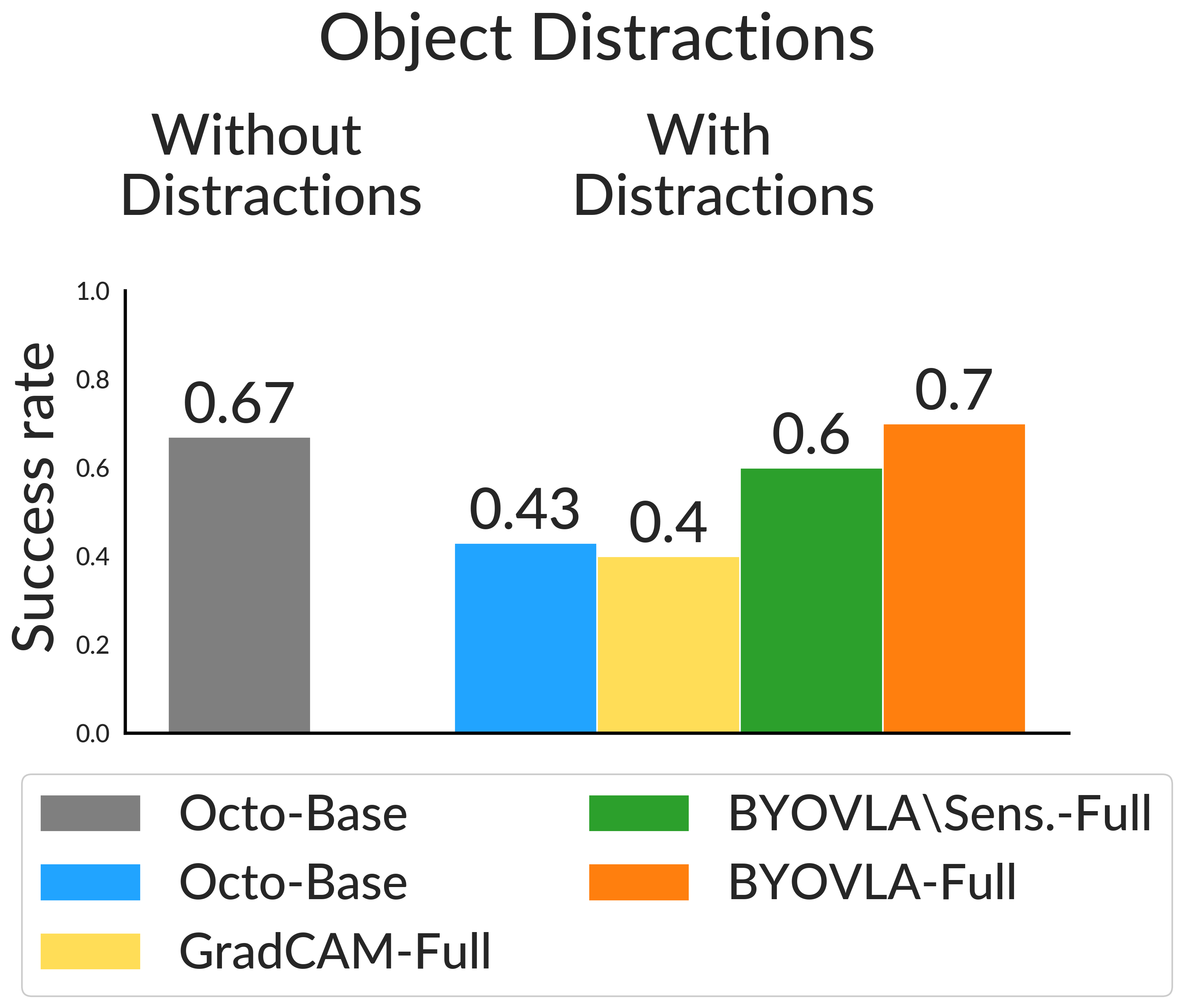}
    \caption{Success rates for \byovla{} with Octo-Base on language instruction "place the carrot on yellow plate" when sampling $K$ observations. }
    \label{fig:octo_K_obs}
\end{figure}

The results are depicted in Fig. \ref{fig:octo_K_obs}, which demonstrates how \byovla{} is still able to achieve the nominal performance of Octo-Base in the presence of distractions. Application of the visual sensitivity probe was applied every $T_a=3$ timesteps, in accordance with the methodology described in Sec. \ref{sec:experiments}. Each method received a total of 30 trials.

Sampling with $K$ observations, instead of $K$ actions, does not appear to yield significantly better task-success rates. For instance, the performance of Octo-Base with \byovla{} is almost unchanged, and \byovla{} $\setminus$ Sens.-Full improved only slightly. We hypothesize this is likely due to experimenting in static environments with simple tasks, and anticipate sampling with $K$ observations will offer benefit in more complex scenarios.

\end{document}